\pdfoutput=1

\documentclass[11pt]{article}

\usepackage[preprint]{neurips_2024}

\usepackage{times}
\usepackage{latexsym}
\usepackage{linguex}
\usepackage{mathtools}
\usepackage{enumitem}
\usepackage{xcolor}
\usepackage{wrapfig}
\usepackage{hyperref}

\usepackage[T1]{fontenc}

\usepackage[utf8]{inputenc}

\usepackage{microtype}

\usepackage{inconsolata}

\definecolor{blue}{HTML}{5698e5}
\definecolor{pink}{HTML}{ee566a}

\title{Missed Causes and Ambiguous Effects: Counterfactuals Pose Challenges\\for Interpreting Neural Networks}

\author{Aaron Mueller \\
Northeastern University \\
Technion -- Israel Institute of Technology \\
\texttt{aa.mueller@northeastern.edu}\\
}

\begin{document}
\maketitle
\begin{abstract}
Interpretability research takes counterfactual theories of causality for granted. Most causal methods rely on counterfactual interventions to inputs or the activations of particular model components, followed by observations of the change in models' output logits or behaviors. While this yields more faithful evidence than correlational methods, counterfactuals nonetheless have key problems that bias our findings in specific and predictable ways. Specifically, (i) counterfactual theories do not effectively capture multiple independently sufficient causes of the same effect, which leads us to miss certain causes entirely; and (ii) counterfactual dependencies in neural networks are generally not transitive, which complicates methods for extracting and interpreting causal graphs from neural networks. We discuss the implications of these challenges for interpretability researchers and propose concrete suggestions for future work.
\end{abstract}

\section{Introduction}

Causal interpretability techniques have become popular, in large part due to the increasing popularity of mechanistic interpretability. Such techniques aim to faithfully understand the \emph{causal mechanisms} underlying an observed model behavior. In practice, ``causal method'' is generally synonymous with ``method that employs counterfactual interventions''; assumptions like these take the \textbf{counterfactual theory of causality} \citep{lewis-1973-counterfactuals} for granted, often assuming its benefits but ignoring its pitfalls. This is problematic: counterfactual theories, while intuitive at first glance, leave us prone to counterintuitive fallacies that can systematically bias our results and interpretations if we are not careful.

In this paper, we will argue that there are two tricky problems that arise when employing counterfactual interventions to interpret neural networks: (i) \textbf{overdetermination}, where multiple simultaneous and similar causes of the same effect will be systematically missed; and (ii) the \textbf{non-transitivity} of counterfactual dependencies, which make it difficult to interpret the relationship between pairs of nodes that are not directly connected. We will start by describing the counterfactual theory of causality (\S\ref{sec:counterfactual_theory}) and its role in interpretability research (\S\ref{sec:counterfacts_in_interp}). Then, we will describe the resulting problems that have permeated interpretability research (\S\ref{sec:nontransitive}, \ref{sec:overdetermination}). Finally, we will discuss the implications of these problems and propose suggestions for future work (\S\ref{sec:discussion}).

\section{The Counterfactual Theory of Causality}\label{sec:counterfactual_theory}
The first counterfactual interpretation of causality was posed by David Hume: ``We may define a cause to be an object\ldots where, if the first object had not been, the second had never existed'' \citep{hume-1748-understanding}. This was later formalized by \citet{lewis-1973-counterfactuals} into what we now recognize as the counterfactual theory. His theory is based on \textbf{counterfactual conditionals}, which take the form ``If cause $C$, then event $E$.'' These conditionals do not necessarily indicate causation (e.g., ``If I were in Berlin, I would be in Germany'' is a valid counterfactual conditional, but the inverse does not hold).

\citet{lewis-1973-counterfactuals} poses that a \textbf{causal dependence} holds iff the following condition holds:

\begin{quote}
``An event $E$ \emph{causally depends} on $C$ [iff] (i) if $C$ had occurred, then $E$ would have occurred, and (ii) if $C$ had not occurred, then $E$ would not have occurred.''
\end{quote}
\citet{lewis-1986-causation} notes that counterfactual dependence is sufficient \emph{but not necessary} for causation; he therefore extends his definition of causal dependence to be whether there is a \emph{causal chain} linking $C$ to $E$. This approach was later extended from a binary notion of whether the effect happens to a more nuanced notion of causes having influence on \emph{how} or \emph{when} events occur \citep{lewis2000influence}. Other work extends notions of cause and effect to continuous measurable quantities \citep{pearl2000causality}, including direct and indirect effects \citep{robins1992indirect,pearl2001effects}; such work generally bases its computations on structural equation modeling \citep{goldberger-1972-structural}.

In the causality literature, a causal chain from $C$ to $E$ is referred to as a \textbf{mechanism}. The field of \textbf{mechanistic interpretability} does not use the term ``mechanistic'' in this way (or in any standard way; \citealp{andreas-2024-interp}), but many of its methods do fit this definition: they aim to locate the causal chain that explains how an input is transformed into a given model behavior.\footnote{We believe this theoretically grounded definition of ``mechanism'' should be adopted in mechanistic interpretability. There do exist non-causal methods which are labeled as mechanistic interpretability, but even if the method does not rely on counterfactuals \emph{per se}, seeking algorithmic understanding is essentially equivalent to seeking human-understandable causal mechanisms that concisely explain model behavior. If so, then even correlational methods can claim to pursue mechanistic understanding, albeit with lower faithfulness and causal efficacy on average than methods which causally implicate units of model computation. A benefit is that this definition may encourage more causally rigorous evaluations to verify claims of mechanistic understanding.}

\subsection{Counterfactual Dependencies in Mechanistic Interpretability}\label{sec:counterfacts_in_interp}
When an interpretability study describes its methods as ``causal'', this typically refers to its use of \textbf{counterfactual interventions}. Here, we will use ``counterfactual'' to refer to any activation that a model component would not naturally take given input $x$ without intervention.\footnote{Some use ``counterfactual'' to refer specifically to inputs where the answer is different from that given the original (often termed ``clean'') input. We believe this usage is too narrow: if a neuron is responsible for some behavior, and one manually sets its activation to some alternate value that flips, nullifies, or otherwise modifies its causal contribution, one will still observe an effect on the (probability of the) behavior in a manner that would not have been possible otherwise. We will use ``patching ablation'' to refer to the form of counterfactual based on inputs that flip the answer, but we will also refer to zero ablations and mean ablations as counterfactuals.} 

In this paradigm, we conceptualize a neural network model $M$ as a causal graph from inputs to outputs, where each model component (e.g., a neuron or an attention head) is a \textbf{mediator} or \textbf{node}, and each connection between components in the computation graph is a causal \textbf{edge}. This follows terminology and modeling practices in \textbf{causal mediation analysis} \citep{pearl2001effects}. Neural networks are amenable to this interpretation because they are \emph{complete} causal graphs that fully explain how any valid input $x$ will be transformed into a corresponding output $y$.

A typical method for implicating a model component in some behavior is \textbf{activation patching} \citep{vig2020causal,finlayson-etal-2021-causal,meng2022locating}. In this method, we first define an input $x_{\text{original}}$ and a corresponding correct completion. If we are using a \textbf{patching ablation}, we also define a minimally different incorrect completion. For example, in subject--verb agreement, if the input $x_{\text{original}}$ is ``The \textbf{\textcolor{blue}{manager}}'', a minimal pair of completions could be $y_\text{original}=$ ``\textbf{\textcolor{blue}{is}}'' (correct) and $y_\text{patch}=$ ``\textbf{\textcolor{pink}{are}}'' (incorrect). One then defines a minimally different input $x_{\text{patch}}$ where the correct and incorrect answer are flipped; in this example, $x_{\text{patch}}$ would be ``The \textbf{\textcolor{pink}{managers}}''.

Given these variables, one runs $x_{\text{patch}}$ through $M$ in a forward pass (denoted $M(x_{\text{patch}})$) and collects the activation $a_{\text{patch}}$ of a model component $a$. Then, one can run $M(x_{\text{original}})$ and collect some output metric that should be sensitive to this component's activation; often, this metric is the logit difference $y=(p_{\text{patch}} - p_{\text{original}}$). Then, one reruns $M(x_{\text{clean}})$, but where $a_{\text{original}}$ is replaced with $a_{\text{patch}}$; i.e., we perform a counterfactual intervention to $a$ by setting its value to what it \emph{would have been} in an alternate input. One then measures how much this intervention changes the target metric $y$.\footnote{There exist other interventions which do not require minimally different patch inputs or binary answers. For example, in the absence of a clear binary incorrect answer, one can also measure the negative logit of the correct completion. Rather than patching in $a_{\text{patch}}$, one could instead set $a$ to its mean activation across a sample of general text (referred to as a \textbf{mean ablation}), or randomly sample an activation from an unrelated input (known as a \textbf{patching ablation}). A \textbf{zero ablation}, where $a$ is replaced with 0, is generally not considered principled, unless one is working with units whose default values are 0 (as when working with sparse autoencoders; see \citealt{bricken2023monosemanticity} and \citealt{cunningham2024sparse}).}

This change in the output metric is the \textbf{indirect effect} (IE; \citealp{pearl2001effects,robins1992indirect}) of $a$ on $y$. 
The IE after activation patching is a common metric in causal interpretability \citep{vig2020causal,finlayson-etal-2021-causal,soulos-etal-2020-discovering}, and has been extended for use in mechanistic interpretability techniques which chain causes into graph structures. Variants include patching via adding Gaussian noise (also termed causal tracing; \citealp{meng2022locating}). One can generalize patching to implicate multiple components simultaneously, as in path patching \citep{wang2023interpretability,goldowskydill2023localizing}. This technique has been applied extensively to understand mechanisms underlying sophisticated language model behaviors \citep{geva-etal-2023-dissecting,yu-etal-2023-characterizing,todd2024function} and to edit language models \citep{meng2022locating,meng2023memit}.

Perhaps the most relevant methods to this work are those used to discover and analyze \textbf{circuits} \citep{wang2023interpretability,conmy2023automated}, which are subgraphs of a neural network that are causally implicated in performing a specific task. These generally employ activation patching, path patching, or some combination thereof. Circuits contain nested dependencies, but even if a node does not have a direct edge to the model output, it is often required that each circuit node directly influence the output when ablated. This may be thought of as a generalization of individual component probing, in that component probing searches for causal graphs of depth $=$ 1, whereas circuit discovery locates causal graphs of depth $>$ 1. Circuits are often used for post-hoc interpretability \citep{hanna2023how,lieberum2023does,chugtai2023toy}, and sometimes to modify model behaviors via targeted ablations \citep{li2024circuit,marks2024sparse}. Another causally grounded line of work takes hypothesized causal graphs as inputs and aligns them with (potentially multidimensional) model subspaces, either via search or direct optimization over intervention site boundaries  \citep{geiger2021causal,wu2023interpretability,geiger2024finding}. Here, patching ablations are referred to as interchange interventions.

\section{Counterfactual dependence is not transitive.}\label{sec:nontransitive}
It is natural to assume that if $C$ counterfactually depends on $B$, and $B$ depends on $A$, then $C$ must depend on $A$. This is a fallacy. Consider the following canonical example from Ned Hall:\footnote{As reported in \citet{hitchcock2001intransitivity} with credit to Hall.}

\ex.\label{ex:hiker}
    \a.[$A$:] A hiker is walking up a mountain. A large boulder begins rolling down the mountain toward the hiker.
    \b.[$B$:] The hiker, noticing the rolling boulder, ducks out of the way.
    \c.[$C$:] The hiker lives.

\begin{wrapfigure}{r}{0.48\linewidth}
    \centering
     \includegraphics[width=\linewidth]{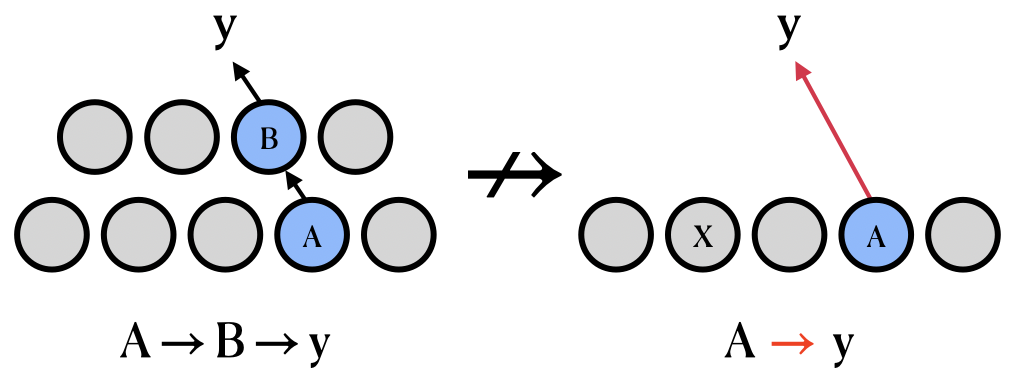}
     \caption{Illustration of the non-transitivity of counterfactual dependence. If an output $y$ depends on $B$, and if $B$ depends on $A$ (potentially in addition to other nodes), then it does not necessarily entail that $y$ depends on $A$.}
     \label{fig:nontransitivity}
\end{wrapfigure}

In this scenario, $C$ depends on $B$, and $B$ depends on $A$. Does it therefore make sense to say that the boulder rolling down the mountain caused the hiker to live? The reason this feels strange is because $C$ is not counterfactually dependent on $A$: the hiker would have lived even if the boulder had never rolled down the mountain. Thus, while each local counterfactual dependence holds, the transitive dependence of $C$ on $A$ does not. In other words, \textbf{counterfactual dependence is not necessarily transitive}. See Figure~\ref{fig:nontransitivity} for an illustration.

However, causality intuitively feels transitive, and there are many cases where it is.\footnote{It has long been debated whether \emph{actual causality}, as separate from counterfactual dependence, is always transitive (cf.\ \citealt{Paul-2013-causality,lewis-1973-counterfactuals,lewis-1986-causation}, who claim that transitivity is a key desideratum of theories of causality). We are inclined to follow the analysis of \citet{Beckers-2017-causality}, who claim that the transitivity of causality is conditional, and who then give sufficient conditions for causal transitivity.}
Consider playing a game of billiards, where one hits billiard ball $A$, which then hits ball $B$, which sinks ball $C$. In this case, $C$ \emph{does} counterfactually depend on $A$. What, then, is the fundamental difference between transitive and non-transitive dependency chains?
\\
\citet{halpern2016sufficient} poses necessary and sufficient conditions to guarantee the transitivity of counterfactual dependence. Suppose there exist random variables $A$, $B$, and $C$. Let there exist values $a_1$ and $a_2$ that $A$ could take, and there exist similar pairs of values $b_1$, $b_2$ and $c_1$, $c_2$ for $B$ and $C$, respectively. Also let ``$\rightarrow$'' indicate entailment. Then the following conditions guarantee the transitive dependence of $C$ on $A$:

\begin{enumerate}[noitemsep,topsep=0pt]
    \item $A=a_1 \rightarrow B=b_1$
    \item $B=b_1 \rightarrow C=c_1$
    \item $c_1 \neq c_2$
    \item $A=a_2 \rightarrow B=b_2$
    \item $A=a_2 \wedge B=b_2 \rightarrow C=c_2$
\end{enumerate}
The intuition is that $A$ can take values which determine at least two distinct values of $B$,\footnote{This may be easier to conceptualize if one treats $a_2$, $b_2$, and $c_2$ as the \emph{default values} of each variable, as \citeauthor{halpern2016sufficient} does in leading up to this more general set of conditions.} and that $B$ can determine at least one value of $C$. Then, if \emph{the disjunction of} $A$ and $B$ also determines a particular value of $C$ that is distinct from the one that $B$ alone caused, then the dependency chain from $A$ to $C$ is transitive, and $C$ therefore counterfactually depends on $A$.

\citeauthor{halpern2016sufficient} poses an additional set of sufficient (but not necessary) conditions to guarantee transitivity:

\begin{enumerate}[noitemsep,topsep=0pt,partopsep=0pt]
    \item For every value $b$ that $B$ can take, there exists a value $a$ that $A$ can take such that $A=a \rightarrow B=b$.
    \item $B$ is in every causal path from $A$ to $C$.
\end{enumerate}
\textbf{In neural networks, counterfactual dependencies will not often be transitive.} While possible, it seems unlikely that either set of conditions would hold in most neural networks. The first is unlikely due to the existence of many neurons that simultaneously influence $B$, such that the probability is low of there existing some value $a_1$ of $A$ that alone causes $B$ to take some value $b_1$. The second is unlikely because, in the simplest case where $B$ is in a hidden layer between $A$ and $C$, there are $n$ possible paths from a neuron in $A$'s layer to a neuron in $C$'s layer, where $n$ is the number of neurons in $B$'s layer. This number grows exponentially as we increase the layer distance between $A$ and $C$. Thus, it is reasonable to assume by default that counterfactual dependencies will generally \emph{not} be transitive in neural networks, unless $B$ is some \textbf{causal bottleneck} through which information from node $A$ \emph{must} pass to arrive at $C$. There is no reason \emph{a priori} to expect such bottlenecks to exist at the level of neurons, attention heads, or sparse autoencoder features, which is unfortunate given that these fine-grained units are more likely to be human-interpretable. One could view residuals as causal bottlenecks into which all information is collapsed at the end of a layer, so causal chains would be transitive if the intermediate node $B$ were the entire residual \emph{vector}; nonetheless, such coarse-grained units are generally difficult to interpret. Future work could consider methods for demonstrating the existence and frequency of interpretable causal bottlenecks in neural networks.

This presents a problematic trade-off. Methods that require components to significantly change the output will not recall components that explain how intermediate causal nodes arise. This hinders our understanding of how a model composes increasingly high-level intermediate abstractions to perform a task. However, if we employ methods which search over \emph{all} local dependencies and do \emph{not} require a significant change to the output, we will include many more nodes than is necessary to explain the model behavior, and probably significantly increase our method's time complexity.

\subsection{Case study: Succession features in Pythia 70M}
Circuits are directed acyclic graphs (DAGs), where each node represents a model component or set of components, and each edge represents a causal link from an upstream node to a downstream node. The leaf node is typically the target metric, such as the logit difference between a correct and a minimally different incorrect completion to some context. Circuits are intended to represent the subgraph of the model's computation graph that is causally implicated in increasing the target metric. To ensure that nodes in the circuit are human-interpretable, one can use sparse autoencoder dictionaries \citep{bricken2023monosemanticity,cunningham2024sparse} to decompose submodules (like MLP or attention layers) into human-interpretable sparse features.

In \citet{marks2024sparse}, the authors discovered sparse feature circuits for various model behaviors, including a \emph{specific succession} behavior where Pythia 70M predicts ``4'' given inputs containing numbered lists (where each number is potentially separated by variable-length text), as in ``1: \ldots\texttt{\textbackslash{}n}2: \ldots\texttt{\textbackslash{}n}3: \ldots\texttt{\textbackslash{}n}''. After generating a sparse feature circuit for this seemingly coherent behavior, they find that it is actually the intersection of two distinct mechanisms: (1) a \emph{general succession} mechanism \citep{gould2024successor}, which promotes the next item in ordinal or numeric lists, and (2) a \emph{specific induction} mechanism \citep{olsson2022context}, which copies previous elements of lists in the same context. As presented, the successor mechanism arises spontaneously at a layer 3 attention feature, which has no in edges; because local dependencies are not considered, \emph{it is unclear how this succession abstraction is composed from the inputs}.

\begin{figure}
    \centering
    \includegraphics[width=\linewidth]{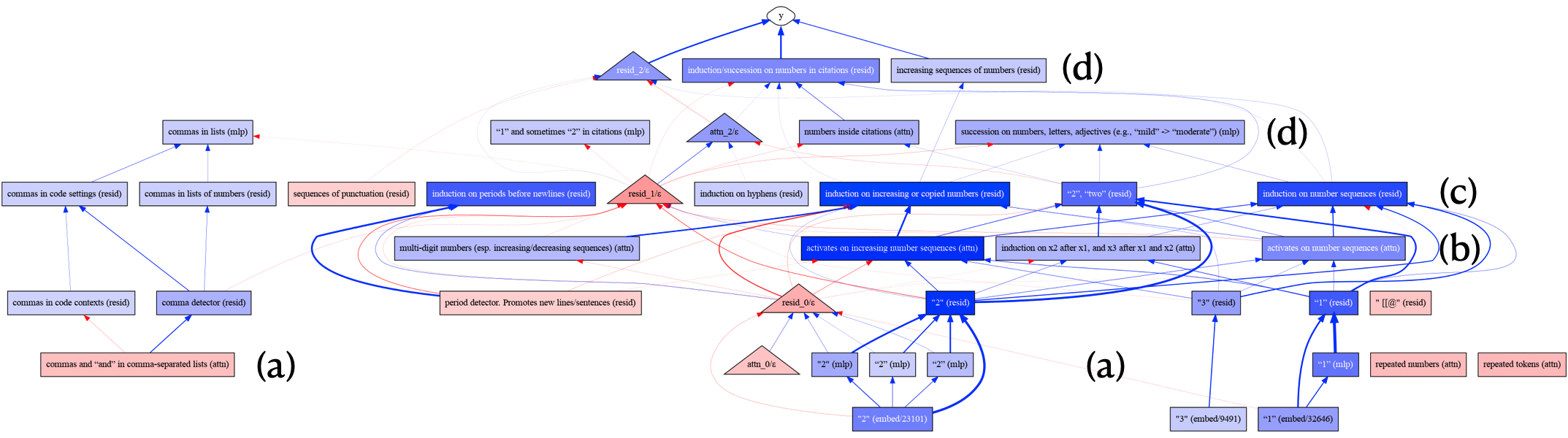}
    \caption{Sparse feature circuit for layer 3 attention feature 14579, a general succession feature (denoted \texttt{y} in the figure). Best viewed zoomed in. Darker blue shades indicate higher-magnitude positive effects, and darker red shades indicate higher-magnitude negative effects. Given inputs such as ``1, 2, 3,'', the model's first layers detect each number and the punctuation between them (a). The number detectors are inputs to number list detectors and increasing number sequence detectors (b). These features then inform induction features (c) that are only active in specific contexts or on particular classes of phenomena (like number sequences). These aid more general incrementor/succession features (d).}
    \label{fig:sfc_attn3_condensed}
\end{figure}

Thus, we obtain a sparse feature circuit (SFC) for cluster 382 from \citet{marks2024sparse}, where the target metric $y$ has been changed from the logit difference of two possible next tokens to the activation of the layer 3 attention succession feature (Figure~\ref{fig:sfc_attn3_condensed}). We use the same method as in \citet{marks2024sparse}, changing only the target metric; see App.~\ref{app:sfc} for details and hyperparameters. We manually annotate each feature by inspecting its activations and the logits that change most when the feature is ablated.\footnote{There are problems in assigning natural language explanations to model components. These are detailed in \citet{huang-etal-2023-rigorously}.} This SFC contains features and edges that were not in the original circuit, but which help explain how succession is computed in Pythia. For example, there is a causal chain from layer-0 features which detect specific numbers, to layer-1 features which detect sequences of numbers (regardless of order), to layer-2 features which detect increasing sequences of numbers. There is another mechanism where layer-0 features activate on all commas, then layer-1 features activate only on commas within lists of numbers, then layer-2 features activate on commas and elements of comma-separated number lists. This yields even stronger evidence that the target attention feature actually is a successor head---but also that it is primarily sensitive to \emph{numeric} succession. 

This is helpful for understanding the original feature circuit: the numeric successor feature that spontaneously arose in the original SFC is actually composed from the intersection of specific number detectors and number list detectors. This anecdote suggests that we should include some purely local dependencies if our goal is to fully understand how higher-level abstractions and decisions are composed from low-level input features. That said, it is easy to see how this would yield an exploding number of features if we include \emph{all possible} local dependencies: the graph would become uninterpretable due to a rapidly increasing description length. For example, in the above SFC, we also found features which activate inside citations of the form ``\texttt{[@<num>]}'', where \texttt{<num>} is an integer. Is this useful? Maybe, but it may reflect how a particular input distribution is handled (this cluster contains a significant number of citations in this format) rather than how a generic behavior is composed. Another framing is that the citation feature is likely to reduce the description length of the feature less than the number list feature would, so we should prioritize explaining the number list features.

Given this discussion, should one keep local dependencies---i.e., upstream nodes that influence downstream nodes which affect the target metric, but not the target metric directly? Or should one instead mandate that each causal node we keep have some direct measurable effect on the target metric? \textbf{If one's goal is to understand how a model will generalize, one should also consider at least \emph{some} local causal dependencies.} However, \textbf{if one's goal is merely to understand which components will directly affect downstream performance} (e.g., when editing or pruning models), \textbf{it may suffice to only include components that directly affect the output}.\footnote{Though \citet{hase2023does} find that causal localizations do not accurately reflect the optimal locations to update parameters.} This is fundamentally a trade-off between minimizing description length and increasing recall of highly explanative components, similar to the trade-off described in \citet{sharkey-2024-sparsify}. In such cases, the best point in the Pareto frontier may simply consist of the minimal number of features needed to understand whether a model is making the right decisions \emph{in the right way}.

To prevent the problem of overly long description lengths when including local dependencies, we recommend to (1) \textbf{use more diverse datasets when discovering and interpreting counterfactual dependencies}, and to (2) \textbf{directly evaluate the out-of-distribution generalization of interpretability methods by using held-out evaluation datasets that are distinct in controlled ways} from the dataset over which we discover dependencies. These will allow us to recover more causally relevant nodes while also helping us filter out those which are too input-specific or high-variance.

\section{Redundant causes are systematically missed.}\label{sec:overdetermination}

Counterfactual theories of causality struggle to handle \textbf{overdetermination}---i.e., cases where an effect has multiple distinct causes which are each sufficient on their own. Consider the following classic example from \citet{Hall2004concepts}:

\ex.\label{ex:rock}
    \a.[$A_1$:] Suzy throws a rock at a glass bottle.
    \b.[$A_2$:] At the same time, Billy throws a rock at the same bottle.
    \c.[$B$:] The rocks shatter the bottle.

If $A_1$ had not occurred, $B$ still would have happened because of $A_2$, and vice versa. Thus, $B$ is not counterfactually dependent on either $A_1$ or $A_2$ in isolation: ablating only one of them would produce no change to $B$. Does this mean that neither rock caused the bottle to break? This is an uncomfortable conclusion.

The above example feels contrived because simultaneous redundant causes of the same event are rare in the real world. However, in neural networks, redundancy is very common. Consider a layer with $n$ neurons in the output vector of the MLP submodule: these are $n$ simultaneous causes, at least some of which are likely to have similar effects on some downstream model component or on the model behavior. Indeed, redundancy is found to be very common in pre-trained neural networks \citep{dalvi-etal-2020-analyzing}, even for abstract phenomena like syntactic agreement \citep{tucker-etal-2022-syntax}; further evidence comes from the fact that models can be distilled \citep{sanh2020distilbert} or pruned \citep{frantar-2023-sparsegpt} without significant loss in performance. This implies that \textbf{current interpretability methods that rely on ablations probably have significantly lower recall than expected, and it is unclear how many or what kinds of components are missing}.

\begin{figure*}[t]
    \centering
    \includegraphics[width=0.7\linewidth]{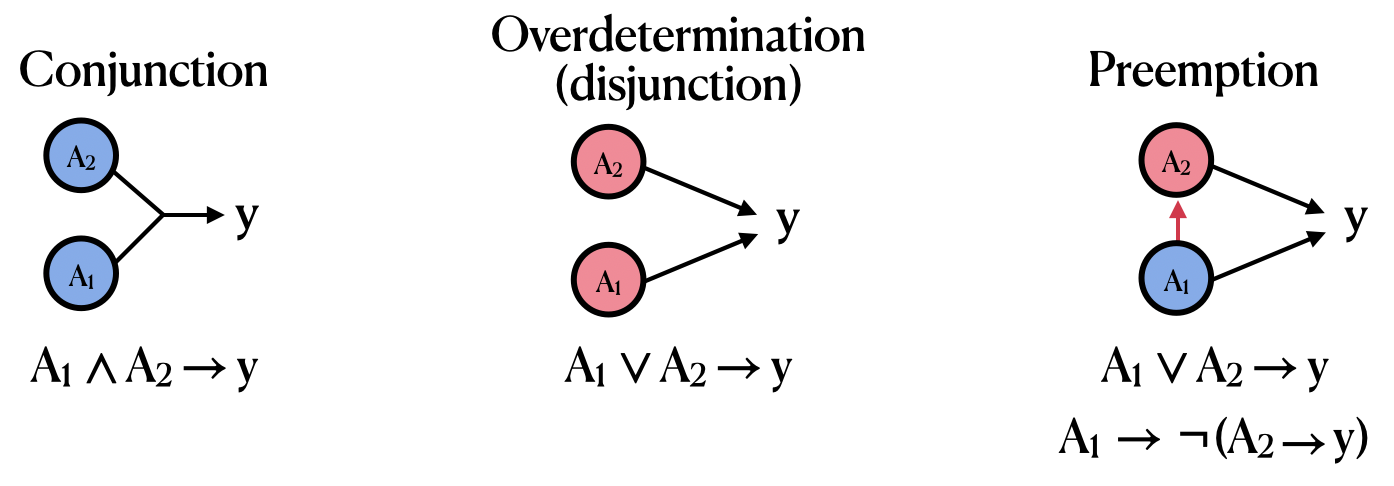}\vspace{-0.5em}
    \caption{Illustration of the result of intervening on individual model components. The captions depict the actual causal graph, while the visualizations depict the causal graph that would have been discovered via ablations to individual components. Nodes in \textbf{\textcolor{blue}{blue}} are in the actual causal graph and would be discoverable. Nodes in \textbf{\textcolor{pink}{pink}} are in the actual causal graph, but would not have been discovered. Black edges are cause-effect relationships, whereas red edges indicate preemption.}
    \label{fig:overdetermination}
\end{figure*}

Suppose for example that we have two neurons $A_1$ and $A_2$ which have activations $a_1$ and $a_2$, respectively, given input $x$. $A_1 = a_1$ and $A_2 = a_2$ cause the activation of $B$ to increase. In the absence of the other, both $A_1=a_1$ and $A_2=a_2$ increase $B$ to some value close to $b$, which causes model behavior $y$. But if both are active, then suppose that the increase in $B$'s activation beyond $b$ is non-linear and small---e.g., it could be logarithmic or at the upper plateau of a logistic function. Thus, in a search over individual model components, intervening on $A_1$ or $A_2$ will have a small marginal effect on $B$. We might therefore be misled into concluding that \emph{neither} $A_1$ \emph{nor} $A_2$ causally contribute to $B$, and thus do not cause the target behavior.
See Figure~\ref{fig:overdetermination} for an illustration. This applies straightforwardly to cases where two neurons have strongly correlated effects, even if their functional roles are not the same. As a less trivial example, assume we are searching for components responsible for successfully performing a question answering task. If $A_1$ detects geography terms, $A_2$ detects political terms, and $B$ predicts country names, then $A_1$ and $A_2$ may have very similar effects on $B$ on certain narrow distributions, despite being sensitive to different phenomena. We might be able to find them if $A_1$ and $A_2$ affect other components or metrics (e.g., examples of opposite labels or from different domains) in distinct and measurable ways. This implies that \textbf{diverse input data and the choice of metric will be crucial to discovering all relevant components}.

There also exist less trivial forms of redundant causation that have already been attested in neural networks. Backup heads \citep{wang2023interpretability} and Hydra effects \citep{mcgrath2023hydra} are examples of what is called \textbf{preemption} in the causality literature. Returning to Ex.~\ref{ex:rock}, if Suzy's rock $A_1$ had hit the bottle first, then Billy's rock $A_2$ would not have shattered the bottle $B$---but preventing $A_1$ would still not prevent $B$ due to $A_2$. See Figure~\ref{fig:overdetermination} for an illustration. In preemption, $A_1$ and $A_2$ are sufficient on their own to cause $y$. Additionally, when $A_1$ is active, it triggers some mechanism that causes $A_2$'s marginal effect on $y$ to be nullified. When $A_1$ is ablated, $A_2$ will cause $y$. Here, we would discover $A_1$ through simple counterfactual ablations, but we would not discover $A_2$ without first ablating $A_1$.

\textbf{Probing individual model components is problematic.} The implications of these phenomena are far-reaching: overdetermination and preemption affect not only circuit discovery methods, but also \emph{any} intervention method where we have many possible simultaneous causes. If one does not test all possible combinations of model components, then one is not guaranteed to recall the full mechanism leading to some model behavior. It may be easier to discover cases of preemption, since one can ablate components that have already been discovered and observe whether new components then affect the output. With overdetermination, it is not clear where to search for these component sets; thus, some type of search over sets of components---rather than individual components---may be necessary.

Given minimally differing input/output pairs, \citet{heimersheim2024use} advocate overcoming this problem without multi-component ablations by using multiple types of patching ablations: one as described in \S\ref{sec:counterfacts_in_interp}, and another where $x_\text{original}$ and $x_\text{patch}$ are swapped (termed ``noising'' and ``denoising'' ablations, respectively). We agree, and to generalize this beyond minimal pairs, we recommend \textbf{using both positive and negative counterfactuals when discovering causal dependencies}---i.e., do not just ablate a component when it is active, but also inject high activations into a component when it is not active. This corresponds to a search where a causal node can be included in the graph if either (i) ablating the node has a negative effect on the behavior, or (ii) injecting it causes the behavior to occur. 

\section{Discussion}\label{sec:discussion}
These challenges pose the following questions for interpretability researchers who employ counterfactuals:

\begin{enumerate}[noitemsep,topsep=0pt]
    \item What should we do with nodes that are locally important, but have little effect on the output?
    \item How can we increase recall without significantly harming precision?
    \item Are counterfactual theories of causality sufficient for interpreting neural networks?
\end{enumerate}

\paragraph{1. Keep at least some local dependencies.}
On (1), one should interpret each edge in a causal graph---such as a circuit edge or a connection between aligned subspaces---as a purely local causal relationship by default.\footnote{Unless all edges represent influence on the output, as in \citet{marks2024sparse}.} One should remain agnostic about the relationship between a given pair of nodes $A$ and $B$ that are not directly connected in the graph---unless the node(s) between them are a causal bottleneck and/or can be determined by $A$, in which case one may assume that $B$ is directly counterfactually dependent on $A$. To ensure that the nodes in a circuit actually affect the outputs, one might respond by testing the effect of a given node on the output metric and only including it if it directly affects the logits. This poses its own problems: just as it did not make sense to talk about the hiker ducking and then surviving without first mentioning the boulder rolling down the mountain in Ex.~\ref{ex:hiker}, it would similarly not make sense to exclude all local causal relationships where a node does not affect the output directly. Indeed, this would be a catastrophic omission when telling the full causal story of how certain intermediate representations which \emph{are} important are composed from simpler abstractions. We therefore advocate to \textbf{keep and analyze components that do not directly affect the outputs by themselves}, but that \emph{do} causally effect other downstream nodes in a circuit.

To prevent the opposite problem of including too many nodes and decreasing precision, we should prioritize including information which is most likely to reduce the description length of the program that implements the target behavior. One way to do this is by \textbf{deploying more diverse datasets}, such that components which only explain a small subset of inputs or minor aspects of the behavior would be less likely to pass our effect size thresholds. We should use input datasets and controlled evaluation datasets which are neither confined to single domains nor highly templatic.

The above discussion focuses on cases where mediators are neurons or attention heads within a single layer. This becomes more confusing when intervening on multiple components: if path patching is used, all nodes within the path should be considered as a single node/mediator. Effects within a path $p_1$ on $y$ will be transitive, but effects between path $p_1$ on the output $y$ via $p_2$ may not be transitive unless $p_1$ and $p_2$ satisfy a set of conditions from \S\ref{sec:nontransitive}. This raises another recommendation: if you employ interventions, \textbf{be explicit about what your causal nodes/mediators are, especially if you are drawing connections between them.}

\paragraph{2. Robust methods will \emph{require} a diversity of interventions and/or interventions to combinations of components.}
On (2), the naïve brute-force solution would be to intervene on all possible combinations of components. Of course, it is not tractable to test all possible combinations of components, even when employing efficient approximations like attribution patching \citep{syed2023attribution,kramar2024atp}. A reasonable compromise may be to rely on greedy search methods (as in \S4.2 of \citealp{vig2020causal}).

More tractable methods for capturing overdetermined causes will probably entail a diversity of data and intervention types. We can, for example, use both positive and negative counterfactuals (as described in \S\ref{sec:overdetermination}). As a starting point, we could construct shared toy models which have overdetermination and preemption built in; using these, we can develop methods that can tractably discover redundant causes. These methods can later be scaled to larger and more capable neural models.

\paragraph{3. Counterfactual theories are sufficient---but we must interpret them correctly.}
On (3): if counterfactual theories do not handle these issues well, perhaps there exist better theories. It would not suffice to return to \textbf{regularity theories} of causality \citep{mill1846system}, which ultimately reduce to equating correlation and causation, and therefore reduce precision via the inclusion of false positives. \textbf{Inferential theories} propose rules under which certain causes are \emph{always} followed by specific effects \citep{mackie1965causes}. There have been significant advances in inferential theories that explicitly tackle overdetermination and non-transitivity \mbox{\citep[e.g.,][]{andreas2021ramsey,andreas2024regularity}}, but despite this, it would be difficult to obtain satisfying empirical evidence of causal relationships as distinct from strong correlations.

An appealing alternative theory could be \textbf{causality as a transfer of conserved quantitities} \citep{Salmon1984-SALSEA,salmon1994causality}: if cause $A$ transfers some conserved quantity to $B$ in the interval of time $(A, B]$, and the same overall quantity remains in $A\ \vee B$ throughout that time, then $A$ has causal influence on $B$. In LSTMs \citep{hochreiter1997lstm} or state space models \citep{gu2021combining,gu2022efficiently}, this quantity could be related to the state vector carried across time steps; causal analyses could therefore be framed in terms of components which write to and erase from this state.\footnote{This is similar to ideas proposed in \citet{ferrando2024information} for Transformers, though they use similarities as a proxy for which components were important to a downstream submodule output, which is not guaranteed to capture faithful causal subgraphs.}
In a Transformer network \citep{vaswani2017attention}, what might this conserved quantity be? There is also \textbf{causality as Kolmogorov complexity reduction} \citep{alexander2023subjective,janzing2010causal}, where the inclusion of cause $C$ reduces the minimum description length of the effect $E$. However, this merely captures variables which sufficiently \emph{describe} the effect, and which do not necessarily \emph{execute} it. Further, this theory has been proposed primarily to capture the human notion of subjective causality, rather than actual causality.

Importantly, many non-counterfactual theories are not necessarily motivated by counterfactuals being inaccurate approximations of causality; rather, many are motivated by counterfactuals being difficult to obtain in observational settings, as in clinical trials. Fortunately, in neural networks, counterfactuals are easy to obtain: we have complete control over computation graphs and full access to the values of and connections between each node in the graph. Thus, despite its challenges, counterfactual theories are still likely the most actionable for interpretability research based on neural networks. Even without access to counterfactual \emph{inputs}, the activation of a particular node can simply be nulled out via setting it to its mean activation, or sampling an activation from an unrelated context. While not fully principled,\footnote{Consider this example from \citet{nanda2023comprehensive}: a pair of neurons' activations may be uniformly distributed at points along a circle; here, the mean activation of the pair would be the center of the circle, which results in activations that the neurons would never actually have on any input.} this is still a tractable way to establish a sufficient condition for causal dependence.

In short, the challenge is not that the counterfactual theory has problems: the challenge is that these problems are difficult---but possible---to tractably address at scale. As a starting point, future work could proceed by creating toy models where overdetermination, preemption, and non-transitive dependencies are built into the network. This would allow us to design methods that can adequately handle the problems outlined here; later, these could be made more scalable via approximate metrics or heuristic-based search methods.

\subsection{We need better causal mediators.}
Much of this paper has assumed that, given access to a causal graph, one can interpret the mediators/nodes in the graph. This is often not the case. Methods where the nodes \emph{can} be interpreted tend to be less flexible and open-ended due to the required effort and compute: for example, sparse feature circuits \citep{marks2024sparse} require significant compute for training autoencoders, and these autoencoders may not generalize to a model that has been significantly tuned, pruned, and/or edited. That said, this is a one-time cost: once these autoencoders have been trained, they can be released and easily deployed for others to use. In distributed alignment search \citep{wu2023interpretability,geiger2024finding}, we are required to have a prior hypothesis as to how a model accomplishes a task, and the evaluation metrics do not tell us in what ways our hypotheses are wrong; thus, refining hypotheses may be non-trivial when a model implements a complex behavior in a non-human-like way.

On the other hand, methods that are more flexible and that do not require prior hypotheses often yield causal graphs with uninterpretable nodes. Thus, more work is needed on producing Pareto improvements with respect to human interpretability and compute-/labor-efficiency. Sparse autoencoder dictionaries represented a push in a more fine-grained direction, but other methods could consider learning more coarse and high-level abstractions---e.g., high-dimensional structures or supersets of components which have a coherent functional role.

\subsection{Possible Counterarguments}
\paragraph{Possible counterargument 1: ``Sacrificing a small amount of recall is acceptable.''} We anticipate one common counterargument to be that sacrificing some recall is acceptable when the search space of mediators (and combinations thereof) is large. We would argue against this as a general recommendation: there may exist classes of phenomena that are more likely to be encoded redundantly than others. This means that systematically excluding overdeterminers may bias our findings in a way that hides particular types of features more than others. Thus, if one aims to obtain the most accurate and complete picture of how a model accomplishes a behavior, then one must capture redundant causes, as well as mediators that primarily affect downstream mediators and not the output metric.

\paragraph{Possible counterargument 2: ``These problems primarily apply to binary notions of causality. Neural networks are continuous.''} Another concern is that the above examples have depended on binary accounts of causality, where an event either causes an effect or it does not. Indeed, neural networks are much more forgiving: they may be interpreted as \emph{probabilistic} or \emph{continuous} causal graphs, where edges between nodes are weighted and where effects may be measured via continuous metrics like indirect effects. Current interpretability methods already rely on such metrics. And indeed, \citet{lewis2000influence} defines causality as whether a cause counterfactually \emph{influences} the manner in which an effect occurs, rather than as a binary notion of whether $E$ happens. This means that causes can have varying degrees and types of influence on a given effect, which makes partial causes easier to handle.

Nonetheless, the same non-transitivity and overdetermination issues still apply. The issue has simply shifted: \emph{indirect effects are not transitively conserved in causal chains}. For example, given a neuron $A$ that strongly influences the activation of neuron $B$, and neuron $B$ that strongly influences output $y$, $A$ may still have a very small effect on $y$. $A$ might trigger or frequently co-occur with another mechanism that nullifies $B$'s effect on $y$. And sets of variables $A_{1,2,\ldots,n}$ that result in $B$ taking similar values would still be difficult to completely recall when using interventions to just one of these variables at a time.

Moreover, causal interpretability methods often define some arbitrary effect size threshold above which we keep some component in a circuit, or where we consider the component to have a significant enough contributition to the output phenomenon to be considered causally relevant. In other words, we start with continuous effect sizes, but then collapse our final results into discrete subsets or subgraphs of components that are considered causes, while excluded components are not considered causes. This effectively gives us binary causal graphs, which makes structures such as circuits even more easily susceptible to the same fallacies described here---at least in their current presentation.

\section{Broader Impact and Limitations}\label{sec:limitations}
This paper discusses counterintuitive challenges that arise under counterfactual theory of causality. These challenges could partially explain why methods that apply or edit discovered components often do not generalize robustly. For example, in model editing, could this explain in part why the most causally relevant components discovered via counterfactual interventions are not the most effective locations to edit \citep{hase2023does}? Additionally, given that information is gradually accumulated across layers \citep{meng2022locating,meng2023memit,geva-etal-2021-transformer,geva-etal-2022-transformer}, might the inclusion of locally but not necessarily globally relevant causal dependencies improve the faithfulness of the causal subgraphs we extract?

We acknowledge that this analysis could be improved or expanded in various ways. While the problems discussed here should impact any method that relies on counterfactual interventions to a subset of neural network components, this paper is written primarily with circuit discovery \citep{elhage2021mathematical,conmy2023automated}, alignment search \citep{geiger2021causal}, or component set discovery methods \citep{vig2020causal} in mind. This is not a representative sample of causal and mechanistic interpretability methods. For example, it is not immediately intuitive how these problems affect methods that derive projections to erase or guard particular concepts in latent space \citep{belrose2023leace,ravfogel-etal-2020-null,elazar-etal-2021-amnesic}, or methods that leverage gradients from parametric classifiers to update the model at specific token positions \citep[e.g.,][]{giulianelli-etal-2018-hood}. One could argue that learned projections or modifications to a single layer will still be susceptible to preemption and non-transitivity concerns, since locally relevant early information and preempted later information may be missed by interventions that only affect the model at a single layer; nonetheless, if this layer is a causal bottleneck, non-transitivity may not apply if the entire vector is modified. This should be explored in more detail in future work.

Additionally, simultaneous overdetermination has not yet been attested in large language models. While there are already examples of preemption \citep{wang2023interpretability,mcgrath2023hydra}, demonstrations that overdetermination is possible in language models \citep{heimersheim2024use}, and demonstrations that redundancy is common \citep{tucker-etal-2022-syntax}, it is not clear whether there exist groups of components in large language models that have the same effect on some downstream component or output behavior when either some subset or all such components in the set are active. Generally, when multiple causes trigger the same effect in a neural network, it will often be \emph{additive}, rather than truly \emph{redundant}: i.e., each component has a measurable marginal contribution to the magnitude of the effect, which makes each component discoverable. Thus, it is still not clear in what settings we would miss simultaneous causes, or in which settings they are most likely to arise.

\section{Conclusion}
We have described two problems for interpretability researchers who base their methods on counterfactual interventions: non-transitivity and overdetermination. The former complicates methods for discovering and interpreting deep causal graphs, while the latter lowers recall in systematic ways. We have discussed the implications of these findings when interpreting findings from mechanistic interpretability studies, and proposed suggestions for future work in this space.

\section*{Acknowledgments}
We are grateful to Eric Todd, Can Rager, Yonatan Belinkov, and David Bau for providing feedback on an earlier version of this work. We are also thankful to the reviewers of the 2024 ICML Workshop on Mechanistic Interpretability for their helpful comments. This work was produced while A.M.\ was funded by a postdoctoral fellowship under the Zuckerman STEM Leadership program.

\bibliography{custom}
\bibliographystyle{iclr2024_conference}

\appendix

\section{Discovering a Sparse Feature Circuit for a Sparse Feature}\label{app:sfc}
In \S\ref{sec:nontransitive}, we discovered a sparse feature circuit containing components that cause a downstream sparse feature to activate at higher positive values. This circuit was discovered with Pythia 70M \citep{biderman2023pythia} using the sparse autoencoder dictionaries released with \citet{marks2024sparse}.\footnote{Released at \url{https://github.com/saprmarks/dictionary_learning}.} We use the code and data released by the authors; the dataset for this succession cluster is from \url{https://feature-circuits.xyz}, and corresponds to Cluster 382 in \citet{marks2024sparse}.

The primary difference between this sparse feature circuit and the original circuit for this cluster is the target metric. In the original work, the metric was defined as the negative log-probability of the token $x_{t+1}$ given some context $x_{1..t}$:

\begin{equation*}
    y = - \log p_\theta(x_{t+1} | x_1,\ldots,x_t)
\end{equation*}

In \S\ref{sec:nontransitive}, the target metric is the negative activation of the layer 3 attention feature at index 14579. Here, we use ``feature'' to refer to a single index of a sparse autoencoder's encoded latent space. Specifically, given the output vector $\mathbf{x}$ of the out projection of the layer 3 attention block, the succession feature is index 14579 of the vector $\mathbf{f}$ defined by the encoder
\begin{equation*}
    \mathbf{f} = \text{ReLU}(W_e(\mathbf{x} - \mathbf{b}_d) + \mathbf{b}_e)
\end{equation*}
This operation represents the encoder of the sparse autoencoder dictionary; $W_e$, $\mathbf{b}_d$, and $\mathbf{b}_e$ are learned terms. Then, we simply set $y = -\mathbf{f}_{14579}$.

A feature is kept in the circuit if its indirect effect (IE) on $y$ surpasses node threshold $T_N$, a hyperparameter. The IE is computed by ablating the activation of upstream sparse feature $\mathbf{a}$, and then measuring the change in $y$; this change should be positive if the node causally contributes to $y$.  In practice, to compute the IE, one must run $O(n)$ forward passes, where $n$ is the number of components; given that there are nearly 600,000 sparse features to search over, this is not tractable nor scalable. Thus, a linear approximation of the indirect effect $\hat{\text{IE}}$ is used, as it requires only two forward passes and one backward pass ($O(1)$ passes) for all model components:

\begin{equation*}
    \hat{\text{IE}} = \frac{\partial y}{\partial \mathbf{a}}\Big|_{\mathbf{a} = \mathbf{a}_\text{original}} \cdot \left(\mathbf{a}_\text{patch} - \mathbf{a}_\text{original}\right)
\end{equation*}
Here, $\frac{\partial y}{\partial \mathbf{a}}\Big|_{\mathbf{a} = \mathbf{a}_\text{original}}$ is the gradient of the sparse feature $\mathbf{a}$ given the original input, whereas $\mathbf{a}_\text{patch}$ and $\mathbf{a}_\text{original}$ are the activation of the feature given the patch input and original input, respectively. For this cluster circuit, there is no patch input; therefore, we set $\mathbf{a}_\text{patch} = 0$, often called a zero ablation. Note that zero ablations, while not principled for neurons (whose default values may not be 0 and where differences in sign are not necessarily meaningful), sparse autoencoders are trained such that any given feature's activation should be zero on most inputs, and only take positive activations when the feature contributes to $\mathbf{x}$ taking different values than its mean. To make this more accurate, a slightly more expensive and more accurate approximation based on integrated gradients \citep{sundararajan2017ig} is used; see \citet{marks2024sparse} for details.

In these experiments, the node threshold $T_N$ is set to 0.4 and the edge threshold $T_E$ is set to 0.04. The circuit was discovered using a single A6000 GPU.

\end{document}